\newif\iffinalversion
\finalversiontrue

\iffinalversion
\newcommand{\change}[1]{#1}
\else
\newcommand{\change}[1]{\textcolor{blue}{#1}}
\fi

\documentclass[letterpaper, 10 pt, journal, twoside]{IEEEtran}

\usepackage{graphics}           
\usepackage{times}              
\usepackage{amsmath}            
\usepackage{amssymb}            
\usepackage{graphicx}
\usepackage{algorithm}
\usepackage{multirow}
\usepackage{url}
\usepackage[noend]{algpseudocode}
\usepackage{booktabs}
\usepackage{color}
\definecolor{instructioncolor}{rgb}{.5,.5,.5}

\usepackage[font=small]{caption}

\def\secref#1{Sec.~\ref{#1}}
\def\figref#1{Fig.~\ref{#1}}
\def\tabref#1{Tab.~\ref{#1}}
\def\eqref#1{Eq.~(\ref{#1})}

\makeatletter
\usepackage{xspace}
\DeclareRobustCommand\onedot{\futurelet\@let@token\@onedot}
\def\@onedot{\ifx\@let@token.\else.\null\fi\xspace}

\def\etal{{et al}\onedot}
\makeatother

\def\etalcite#1{\etal~\cite{#1}}

\usepackage{array}
\newcolumntype{L}[1]{>{\raggedright\let\newline\\\arraybackslash\hspace{0pt}}m{#1}}
\newcolumntype{C}[1]{>{\centering\let\newline\\\arraybackslash\hspace{0pt}}m{#1}}
\newcolumntype{R}[1]{>{\raggedleft\let\newline\\\arraybackslash\hspace{0pt}}m{#1}}

\newcommand{\RR}{\mathbb{R}}

\usepackage{hyperref}
\usepackage{multirow}
\usepackage{graphicx}
\usepackage{soul}

\usepackage[table,xcdraw]{xcolor}
\title{ Gaussian Radar Transformer\\ for Semantic Segmentation in Noisy Radar Data}

\author{Matthias Zeller$^{1}$\qquad \and Jens Behley$^{2}$\qquad \and Michael Heidingsfeld$^{3}$\qquad \and Cyrill Stachniss$^{4}$%

  \thanks{Manuscript received: July 5, 2022; Revised: Sept 29, 2022; Accepted: Nov 17, 2022. 
  This paper was recommended for publication by Editor Markus Vincze upon evaluation of the Associate Editor and Reviewers' comments.}%
    \thanks{$^{1}$Matthias Zeller is with CARIAD SE and with the University of Bonn, Germany. $^{2}$Jens Behley is with the University of Bonn, Germany. $^{3}$Michael Heidingsfeld is with CARIAD SE, Germany. $^{4}$Cyrill Stachniss is with the University of Bonn, Germany, with the Department of Engineering Science at the University of Oxford, UK, and with the Lamarr Institute for Machine Learning and Artificial Intelligence, Germany}%
\thanks{This is a preprint of the article accepted at Robotics and Automation
Letters (RA-L). © 2022 IEEE.}
  \thanks{Digital Object Identifier (DOI) 10.1109/LRA.2022.3226030}

}

\begin{document}
\maketitle

\markboth{IEEE Robotics and Automation Letters. Preprint Version. Accepted November, 2022.}
{Zeller \MakeLowercase{\textit{et al.}}: Gaussian Radar Transformer for Semantic Segmentation in Noisy Radar Data}

\begin{abstract}
Scene understanding is crucial for autonomous robots in dynamic environments for making future state predictions, avoiding collisions, and path planning. Camera and LiDAR perception made tremendous progress in recent years, but face limitations under adverse weather conditions. To leverage the full potential of multi-modal sensor suites, radar sensors are essential for safety critical tasks and are already installed in most new vehicles today.
In this paper, we address the problem of semantic segmentation of moving objects in radar point clouds to enhance the perception of the environment with another sensor modality.
Instead of aggregating multiple scans to densify the point clouds, we propose a novel approach based on the self-attention mechanism to accurately perform sparse, single-scan segmentation.
Our approach, called Gaussian Radar Transformer, includes the newly introduced Gaussian transformer layer, which replaces the softmax normalization by a Gaussian function to decouple the contribution of individual points. To tackle the challenge of the transformer to capture long-range dependencies, we propose our attentive up- and downsampling modules to enlarge the receptive field and capture strong spatial relations. 
We compare our approach to other state-of-the-art methods on the RadarScenes data set and show superior segmentation quality in diverse environments, even without exploiting temporal information. 
\end{abstract}


\section{Introduction}
\label{sec:intro}
\IEEEPARstart{A}{utonomous} vehicles need to understand their surroundings to safely navigate in dynamic, real-world environments. To achieve holistic perception \change{and enhance safety}, the sensor suites of autonomous vehicles are versatile to \change{explore redundant information} of individual sensors such as cameras, LiDAR, or radar. Particularly in autonomous driving, where a malfunction of one modality can result in lethal consequences, \change{redundancy is key}. Widely explored cameras and LiDAR sensors capture the environment precisely but face limitations under adverse weather such as fog, rain, and snow. Additional information is required, which is accessible via radar sensors, and hence, makes them crucial to enable safe autonomous mobility. 

\begin{figure}[t]
  \centering
  \fontsize{10pt}{10pt}\selectfont
     \def\svgwidth{\linewidth}
     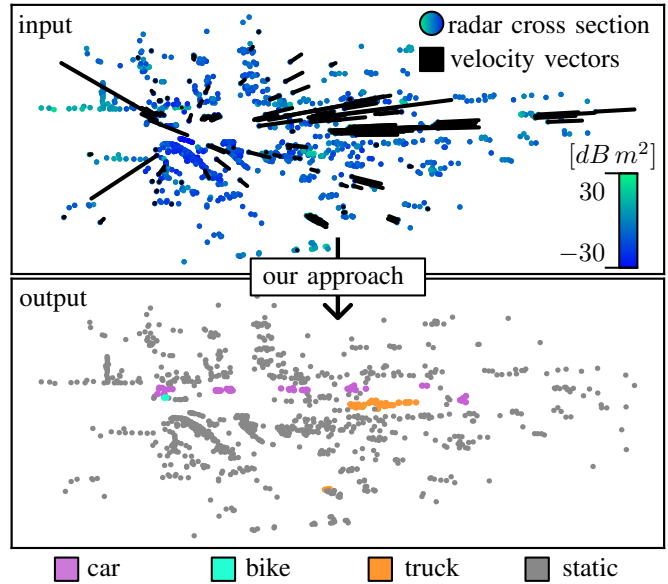
  \caption{Our method performs semantic segmentation of moving objects (bottom) from 4D sparse, single-scan radar point clouds (top) exploiting additional information including the velocity and the radar cross section. In the bottom image, each color represents a different semantic class for moving objects (static is grey).}
  \label{fig:motivation}
  \vspace{-0.6cm}
\end{figure}

In this work, we investigate the semantic segmentation of moving objects in radar point clouds. This task requires differentiating between detections of moving and static objects and assigning a class label to each radar detection, as illustrated in \figref{fig:motivation}. Compared to LiDAR point clouds, radar point clouds are noisier due to sensor noise and multi-path propagation and more sparse. However, radar sensors provide additional information such as the relative velocity to directly indicate moving objects, making the sensor inherently suitable for single-scan processing. Furthermore, the radar cross section \change{depends on} the structure, material, and surface of the reflections, \change{which helps} to differentiate objects.

Most state-of-the-art methods for estimating semantics from radar data~\cite{scheiner2019cvpr,schumann2020tiv} strongly rely on the aggregation of information over multiple scans to accurately perform semantic segmentation. However, aggregation inherently introduces latency, making it unsuitable for tasks requiring immediate information about the vehicle's vicinity, such as collision avoidance. Therefore, this work investigates the processing of single scans by exploiting the additional information provided by radar sensors.

The main contribution of this paper is a new method for accurate, single-scan, radar-only semantic segmentation of moving objects. It takes sparse point cloud representations of radar scans as input and outputs a semantic label for each point. To extract discriminative point-wise features, we build on the self-attention mechanism, a fully attentive neural network with our novel Gaussian transformer layer, and our attentive up- and downsampling modules as central building blocks. We optimize the transformer layer and enable the decoupling via the usage of a Gaussian. Furthermore, our attentive sampling enables the capturing of complex local structures and progressively increases the receptive field of individual points. We combine these building blocks in our new backbone, called Gaussian radar transformer, to enhance feature extraction on sparse and noisy radar point clouds.

In sum, we make three key claims: Firstly, our approach demonstrates state-of-the-art performance for semantic segmentation of moving objects in sparse, single-scan radar point clouds without aggregating multiple scans and without exploiting temporal dependencies. Secondly, the Gaussian transformer layer and the attentive up-and downsampling modules improve feature extraction by decoupling individual points and enlarging the receptive field to enhance accuracy. Thirdly, our fully attentive network is able to \change{extract discriminable features from additional sensor information such as Doppler velocity and radar cross section.}
\begin{figure*}[t]
 \centering
 \fontsize{8pt}{8pt}\selectfont
 \def\svgwidth{\textwidth}
 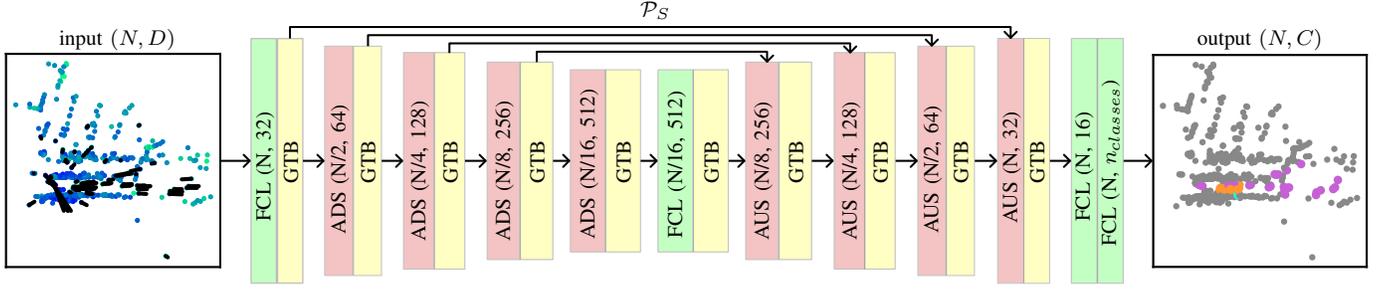
 \vspace{-0.1cm}
 \caption{The architecture of our Gaussian Radar Transformer for semantic segmentation of moving objects. FCL: fully connected layer, ADS: attentive downsampling, AUS: attentive upsampling, GTB: Gaussian transformer block}
 \label{fig:architecture}
 \vspace{-0.5cm}
\end{figure*}
\section{Related Work}
\label{sec:related}

There is extensive literature on semantic segmentation of point clouds, mostly, however, working on LiDAR data. The works can be categorized into projection-based, voxel-based, point-based, and hybrid methods~\cite{guo2021pami}. 

\textbf{Projection-based} methods are inspired by the successful convolutional neural networks~(CNNs)~\cite{lecun1995ann,milioto2019icra}. SqueezeSeg~\cite{wu2018icra}, SqueezeSegV2~\change{\cite{wu2019icran}}, RangeNet++~\cite{milioto2019iros}, and SalsaNext~\cite{cortinhal2020svc} project the point cloud into frontal view images or 2D range images to exploit 2D convolutions. Milioto~\etalcite{milioto2019iros} further alleviate the problem of blurry CNN output and discretization errors by efficient GPU-enabled projective nearest neighbor search as a post-processing step to enhance segmentation results. However, projection-based methods face several problems due to intermediate representation including discretization errors and occlusion. 

\textbf{Voxel-based.} To maintain the 3D geometric information between the data points, voxel-based encoding can be used. VoxSegNet~\cite{wang2019vcg} voxelizes the point clouds as dense cuboids and leverages atrous 3D convolution and attention-based aggregation to enhance feature extraction under limited resolution. Since outdoor point clouds are sparse and vary in density, just a small percentage of voxels are occupied. This makes it inefficient to apply dense convolution neural networks. To reduce the computational burden, Graham~\etalcite{graham2018cvpr} propose submanifold sparse convolutional networks which only generate outputs for occupied voxels. 
Following Polarnet~\cite{zhang2020cvpr}, Zhu~\etalcite{zhu2021cvpr} introduce the cylindrical partitioning, which does not alter the 3D topology compared to the 2D approach, and processes the features by asymmetrical 3D convolution networks.
The advancement in 3D point cloud processing has led to state-of-the-art results of $(\text{AF})^2\text{-S3Net}$~\cite{cheng2021cvpr} and RPVNet~\cite{xu2021iccv} in the SemanticKITTI LiDAR point cloud semantic segmentation benchmark~\cite{behley2019iccv}. Xu~\etalcite{xu2021iccv} combine the voxel-based method with point- and projection-based encoding, utilizing a gated fusion module to adaptively merge the features leading to a hybrid approach.
Since voxel-based methods inherently introduce discretization artifacts and information loss, the hybrid method utilizes point-wise information to alleviate the lossy encoding of information.

\textbf{Point-based.} To leverage the full potential of 3D points, especially for sparse point clouds, and keep the geometric information intact, point-based methods~\change{\cite{landrieu2018cvpr,qi2017cvpr,thomas2019iccv}} have been introduced. %
The pioneering work of Qi~\etalcite{qi2017cvpr} consumes point clouds directly by shared multi-layer perceptrons~(MLPs) and aggregates nearby information by symmetrical pooling functions. The successor PointNet++~\cite{qi2017nips2} groups points hierarchically and progressively extracts features from larger local regions. Schumann~\etalcite{schumann2018icif} adapt the approach and optimize the network for sparse radar point \change{cloud processing}. However, the ability to capture local 3D structures is limited, especially in sparse point clouds. To circumvent, Schumann~\etalcite{schumann2020tiv} aggregate scans, include additional features, or exploit strong temporal relationships. 
To combine local features and reduce the computational cost point-based methods benefit from effective sampling strategies~\cite{hu2020cvpr,wu2019cvpr,yang2020ijcv,yang2019cvpr}. The most frequently used methods for small-scale point clouds are farthest point sampling~\cite{qi2017nips2} and inverse density sampling~\cite{wu2019cvpr}.

Another approach to learning per-point local features is kernel-based convolutions. PointConv~\cite{wu2019cvpr} uses an MLP whereas \change{KPConv}~\cite{thomas2019iccv} defines an explicit convolution to directly learn the kernel. \change{Nobis~\etalcite{nobis2021as} extended \mbox{KPConv~\cite{thomas2019iccv}} and exploit the time dimension of multiple radar scans to perform object detection.} 
Another method to elaborate a stronger connection of the individual points is graph-based, conducting message passing on the constructed graphs~\cite{landrieu2018cvpr}. PointWeb~\cite{zhao2019cvpr2} uses adaptive feature adjustment to represent regions and capture local interactions. However, graph-based networks capture edge relationships of local patches which are invariant to the deformation of these. Velickovic~\etalcite{velickovic2017iclr} and Wang~\etalcite{wang2019cvpr} utilize the self-attention mechanism which is inherently permutation invariant to leverage the limitations and further improve the accuracy.

Self-attention models have revolutionized natural language processing~\change{\cite{dai2019acl,vaswani2017nips}} and inspired self-attention modules for image recognition~\change{\cite{kolesnikov2021iclr,ramachandran2019nips,zhao2020cvpr}} and point cloud processing~\cite{xie2018cvpr,yang2019cvpr}. Recent point transformer networks~\change{\cite{guo2021pct,lai2022cvpr,zhang2021arxiv,zhao2021iccv}} enhance state-of-the-art performance for 3D point cloud understanding by elaborating the self-attention mechanism. PCT~\cite{guo2021pct} proposes offset-attention to sharpen the attention weights by element-wise subtraction of the self-attention features and the input features. Point Transformer uses the vector-based subtraction attention~\cite{zhao2020cvpr} to aggregate local features whereas Stratified Transformer applies dot-product attention and increases the effective receptive field by a window-based key-sampling strategy. Furthermore, recent work elaborates positional encoding to enhance accuracy and keep position information throughout the network~\cite{li2021nips}. 

In contrast to the related work, we propose a novel architecture inspired by self-attention and point transformers. With our newly introduced Gaussian Radar Transformer we are able to capture complex structures in sparse point clouds and further extend the capabilities of the self-attention mechanism. Furthermore, our proposed fully attentive network includes advanced sampling strategies and substantially enhances state-of-the-art performance for semantic segmentation of moving objects in radar point clouds.

\section{Our Approach}
\label{sec:main}
The goal of our approach is to achieve accurate semantic segmentation of moving objects in single-scan, sparse radar point clouds to enhance scene understanding of autonomous vehicles. To accomplish this, we introduce a point-based framework \change{to} directly processes the input point cloud to omit information loss\change{,} and builds upon the successful self-attention mechanism throughout the network. \figref{fig:architecture} depicts our Gaussian Radar Transformer~(GRT)\@. We \change{adopt the encoder-decoder structure of the Point Transformer~\cite{zhao2021iccv}. We replace each module and use our} Gaussian transformer layer as the central building block of each stage, which enables decoupled fine-grained feature aggregation. Furthermore, we introduce attentive up- and downsampling modules to enlarge the receptive field and extract discriminative features.
\subsection{Transformers}
\label{sec:Transformer}
Before presenting our contribution, we shortly revisit transformers as they are a key ingredient in our work. Transformers and self-attention networks rely on the encoded representation of the input features $\mathbf{x}^F \in \RR^{D}$ within the queries $\mathbf{q}$, the keys $\mathbf{k}$, and the values $\mathbf{v}$, as follows:
\begin{align}
\mathbf{q} &= \mathbf{W}_q \mathbf{x}^F, &
\mathbf{k} &= \mathbf{W}_k \mathbf{x}^F, &
\mathbf{v} &= \mathbf{W}_v \mathbf{x}^F,
\end{align}
where $ \mathbf{W}_{q} \in \RR^{D\times D}$, $ \mathbf{W}_{k} \in \RR^{D\times D}$ and $ \mathbf{W}_{v} \in \RR^{D\times D}$ are the corresponding learned matrices of fully connected layers or multi-layer perceptrons~(MLPs). To calculate the attention scores \change{$\mathbf{A}_{i,j}$}, different methods exist such as scalar dot-product~\cite{vaswani2017nips} and vector attention~\cite{zhao2020cvpr}. The \change{scaling by the factor $d_C$ is intended} to counteract the effect of small gradients for the softmax if it grows large in magnitude and is defined as follows:
\begin{align}
\change{\mathbf{A}_{i,j}= \text{softmax}\left(\frac{\mathbf{q}_{i} \mathbf{k}^{\top}_{j}}{\sqrt{d_C}}\right)}.
\label{eq:3}
\end{align}

\change{There is an alternative way for the} weighting of individual feature channels \change{by vector attention that} utilizes relation functions~$f$ such as addition or subtraction. To keep fine-grained position information throughout the network, Wu~\etalcite{wu2021iccv} and Zhao~\etalcite{zhao2021iccv} use relative positional encoding \change{\mbox{$\mathbf{r}_{i,j}=\mathbf{p}_i-\mathbf{p}_j, 1\leq i,j \leq N_l$}}. The final attention weights \change{$\mathbf{A}_{i,j}$} are determined by the softmax function:
\begin{align}
\change{\mathbf{A}_{i,j}}= \text{softmax}(f(\mathbf{q}_{i},\mathbf{k}_{j})+\mathbf{r}_{i,j}).
\label{eq:4}
\end{align}

Since global self-attention leads to unacceptable memory consumption and computational cost the inputs are restricted to local areas with $N_l$ points determined by farthest point sampling and $k$ nearest neighbor~($k$NN)~\change{\cite{qi2017nips2,zhao2021iccv}}. 
The intermediate representation $\mathbf{y}_j$ utilizing vector attention is calculated as follows:
\begin{align}
\mathbf{y}_j &= \sum_{i=1}^{N_l}{\change{\mathbf{A}_{i,j} \odot \mathbf{v}_i }}.
\label{eq:42}
\end{align}

The aggregated features $\mathbf{y}$ are processed by an MLP with a learnable weight matrix $\mathbf{W}_{y} \in \RR^{D\times D}$: 
\begin{align}
\mathbf{o} &= \mathbf{W}_y \mathbf{y}, 
\label{eq:5}
\end{align}
to calculate the final output $\mathbf{o}$.

\begin{figure*}[t]
 \centering
 \fontsize{8pt}{8pt}\selectfont
 \def\svgwidth{0.95\textwidth}
 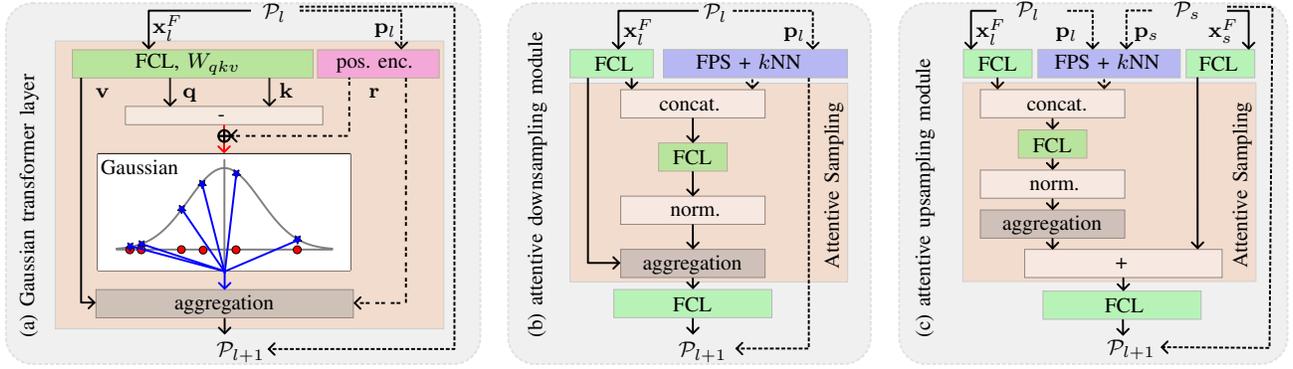
 \caption{The detailed design of each module of our Gaussian Radar Transformer (a) shows the Gaussian transformer layer, (b) the attentive downsampling module, and (c) the attentive upsampling module. FCL: fully connected layer, pos. enc.: positional encoding, concat.: concatenation, norm.: normalization, FPS: farthest point sampling, $k$NN: $k$ nearest neighbor}
 \vspace{-0.1cm}
 \label{fig:modules}
  \vspace{-0.5cm}
\end{figure*}
\subsection{Gaussian Transformer Layer}
\label{sec:gtl}
In sparse radar point clouds, individual reflections contain essential information for downstream tasks such as semantic segmentation of moving objects. To enable independent and precise feature aggregation, we introduce the Gaussian transformer layer~(GTL) \change{based on the Point Transformer layer~\cite{zhao2021iccv}} including vector self-attention as illustrated in \figref{fig:modules}~(a).
\change{Contrary to other approaches, including the Point Transformer~\cite{zhao2021iccv}, which focuses on dense point clouds, we do not utilize the softmax function}, which is defined as:
\begin{align}
   s_{i} = \frac{\exp(z_{i})}{\sum_{j=1}^{N_l} \exp(z_{j})}.
\end{align}

The softmax function leads to a coupling of points since individual outputs $s_i$ are dependent on all inputs $z_j$ with \change{\mbox{$j \in \{1, \dots, N_l\}$}}, which is why the softmax function is also not scale invariant (weighting for dot-product attention \secref{sec:Transformer}).
Furthermore, the backpropagation of the loss $\mathcal{L}$ through the softmax function to obtain the partial derivative $\frac{\partial \mathcal{L}}{\partial z_j}$ to determine the gradients at the input is dependent on all output values. The calculation of the chain rule of derivatives for the softmax can be expressed by the Jacobian matrix $ \mathbf{J}_{\text{softmax}}$ as follows: 
\begin{align}
     \frac{\partial  \mathcal{L}}{\partial z}=  \mathbf{J}_{\text{softmax}}\frac{\partial  \mathcal{L}}{\partial s}.  
\end{align}

If the output values grow in magnitude the gradients diminish since the Jacobian converges to a zero matrix. Hence, the error propagation is restricted, which slows down the learning process.
In contrast, we argue that points belonging to the same class should aggregate the information, whereas points belonging to different classes reduce the information aggregation to a minimum, both of which can lead to a close to zero Jacobian matrix. To overcome this limitation, we replace the softmax function \change{in~\eqref{eq:4} by a Gaussian function $G$, which is executed on every dimension of the vector}, for vector self-attention: 
\begin{align}
\change{\mathbf{A}_{i,j}}= G(f(\mathbf{q}_{i},\mathbf{k}_{j})+\mathbf{r}_{i,j}),
\end{align}
to assess fine-grained information flow for sparse radar point clouds. 
Since the Gaussian function is applied to each feature individually, the points are decoupled, which enables a precise information aggregation to enhance feature extraction and performance. Moreover, the partial derivative of the Gaussian depends on a single output value $s_j$. Hence, vanishing gradients may influence individual points but not whole local areas, which can be seen by the chain rule: 
\begin{align}
     \frac{\partial  \mathcal{L}}{\partial z_j}= \frac{\partial  \mathcal{L}}{\partial s_j} \frac{\partial  s_j}{\partial z_j}.
\end{align}

To derive the output $\mathbf{o}$, we calculate the sum of the
element-wise multiplication:
\begin{align}
\mathbf{o}_j &= \sum_{i=1}^{N_l}{ \change{\mathbf{A}_{i,j}\odot \mathbf{v}_i }},
\end{align}
without further processing by a linear layer reducing computational cost in contrast to \eqref{eq:5}. \change{Following Qi~\etalcite{qi2017nips2} and Zhao~\etalcite{zhao2021iccv}}, we determine the local areas by farthest point sampling and $k$NN algorithm with $k=N_l$. We directly derive the queries $\mathbf{q}_{i}$, the keys $\mathbf{k}_{i}$, and the values $\mathbf{v}_{i}$ by applying a fully connected layer with weight matrix \change{\mbox{$\mathbf{W}_{qkv} \in \RR^{D\times 3D}$}}. For the positional encoding, we adopt the approach of Zhao~\etalcite{zhao2021iccv}. We process the relative position by two fully connected layers and \change{replace the activation function by the} Gaussian error linear unit~(GELU)~\cite{hendrycks2016corr} to determine the positional encoding.

\subsection{Gaussian Transformer Block}
\label{sec:gtb}
Our Gaussian transformer layer~(GTL) is embedded into the center of the Gaussian transformer block~(GTB) which is a residual block, \change{similar to the Point Transformer block~\cite{zhao2021iccv}}, with two fully connected layers processing the input and the output. \change{We replace the activation function with GELU after each fully connected layer}. The GTB processes point clouds $\mathcal{P}$ with point coordinates \change{\mbox{$\mathbf{p}_i \in \RR^{2}$}} and point-wise features ($\mathcal{X}^F=\{\mathbf{x}^F_1,\dots,\mathbf{x}^F_{N}\}$), where $\mathbf{x}^F_i\in\RR^{D}$ with feature dimension~$D$. The features of the individual points $\mathbf{x}^F_i$ are enriched by the information aggregation within the block enhanced by the GTL\@. The point coordinates $\mathbf{p}_i$ are utilized to calculate the positional encoding but not further transformed to keep detailed position information.

\subsection{Attentive Downsampling Layer}
\label{sec:atdown}
To reduce the cardinality of the point cloud $\mathcal{P}_{l+1} \subset \mathcal{P}_{l}$ and thereby the number of points $N$, we process the point cloud by the attentive downsampling layer, depicted in \figref{fig:modules}~(b). Our approach aims to enable adequate sampling and feature processing by applying the self-attention mechanism throughout the network. To reduce computational complexity, we follow Yang~\etalcite{yang2020ijcv} and calculate the attention weights by a single feed-forward layer with the weight matrix $\mathbf{W}_f \in \RR^{(D+2)\times D}$ and no direct representation of keys, queries, and values. We concatenate the input features $\mathbf{x}^F_i$ and the point coordinates $\mathbf{p}_i$ to include positional information to calculate the attention weights \change{$\mathbf{A}_{i,j}$}. Additionally, we normalize the attention weights over the whole point cloud to amplify the contribution of valuable points. The final weights are multiplied with the input features $\mathbf{x}^F$ within local areas which are determined by farthest point sampling and the $k$NN algorithm~\change{\cite{qi2017nips2}}, with $k=N_d$ resulting in:
\begin{align}
\mathbf{y}_i &= \sum_{j=1}^{N_d}{\change{\mathbf{A}_{i,j} \odot \mathbf{x}^F_j}}.
\end{align}

The features are fed into another fully connected layer with LayerNorm~\cite{xiong2020icml} and a GELU activation function. \change{In contrast to Point Transformer~\cite{zhao2021iccv}, which utilizes farthest point sampling and max pooling~\cite{qi2017nips2}, our attentive downsampling includes the information of nearby points, which we assume as valuable for sparse point clouds.}

\subsection{Attentive Upsampling Layer}
\label{sec:atup}
To deduce discriminative features, we argue that the upsampling and feature concatenation of the skip connection is crucial to further enhance performance. The common method for upsampling, \change{also utilized by Point Transformer~\cite{zhao2021iccv},} is an interpolation of the $k=3$ nearest neighbors based on an inverse distance weighted average~\cite{qi2017nips2}. The interpolated points $N_u$ are concatenated with the features of the points, which are passed through the skip connection. The inverse distance weighted average does not include further feature-based information. Hence, the interpolation combines the features only based on their relative position. This is reasonable for dense point clouds because nearby points often belong to the same class. 

However, this might be problematic for sparse point clouds, especially for small instances, which are represented by single points. Therefore, we consider upsampling as an important part to improve feature extraction and propose the attentive upsampling layer. The upsampling layer, which is illustrated in \figref{fig:modules}~(c), first processes the features of the skip connection and the proceeding GTB by two separate fully connected layers with LayerNorm and GELU activation function. To propagate the points from $\mathcal{P}_{l} $ to $\mathcal{P}_{l+1}$ where $\mathcal{P}_{l} \subset \mathcal{P}_{l+1} $ with $N_l \leq N_{l+1}$, we feed the position information of the two point sets and the corresponding features into our attentive upsampling layer. We calculate the $k$ nearest neighbors of the individual points for the point set of the skip connection $\mathcal{P}_{s}$ within the point cloud which has to be upsampled $\mathcal{P}_{l}$. The attention mechanism enables information aggregation of larger local areas since the attention weights will control the information flow and not reduce the discriminability which is possible if large local regions are interpolated. To integrate the positional information we calculate the relative position of the $k$NN of the two point sets given by:
\begin{align}
\mathbf{r}_{i,j}=\mathbf{p}_i-\mathbf{p}_j, 
\end{align}
where $\mathbf{p}_j \in \mathcal{P}_{s}$ and $\mathbf{p}_i \in \mathcal{P}_{l}$. The relative distances $\mathbf{r}_{i,j}$ are concatenated with the features. Following our downsampling layer, we calculate the attention weights directly by processing the concatenated features with a fully connected layer and normalizing the weights over the whole point cloud. The output of the summation is processed by a fully connected layer with LayerNorm and GELU activation function. The self-attention mechanism turns into an inter-attention between the two point clouds to enable attentive feature aggregation. The upsampling is repeated until we have broadcasted the features to the original set of points. We optimize the information aggregation by determining the weighting based on the relative position and the features. We emphasize that the sampling steps are essential for appropriate feature extraction of transformer architectures for sparse point clouds. 

\subsection{Input Features }
\label{sec:inputfeat}
The input is a sparse radar point cloud with $N$ points, feature dimension $D$, and batch size $b$. Each point $\mathbf{p}_{i}$ is defined by two spatial coordinates $x_{i}$, $y_{i}$. Additionally, the radar sensors provide the ego-motion compensated Doppler velocity $v_i$ and the radar cross section $\sigma_i$ resulting in a \mbox{4-dimensional} input vector \change{$\mathbf{x}^F_i=(x_i, y_i, v_i, \sigma_i)^{\top}$}.
\begin{table*}[t]
\resizebox{\textwidth}{!}{
\begin{tabular}{lccc|cccccc|cccccc}
\hline
                           & \multicolumn{1}{l}{}                                  &                                &                                & \multicolumn{6}{c|}{IoU}                                                                                                                                                                            & \multicolumn{6}{c}{F1}                                                                                                                                                                              \\ \cline{5-16} 
\multirow{-2}{*}{Method}   & \multicolumn{1}{l}{\multirow{-2}{*}{Input}}           & \multirow{-2}{*}{mIoU}         & \multirow{-2}{*}{F1}           & static                         & car                            & ped.                           & ped. grp.                      & bike                           & truck                          & static                         & car                            & ped.                           & ped. grp.                      & bike                           & truck                          \\ \hline
\rowcolor[HTML]{E0E0E0} 
RadarPNv1~\cite{schumann2018icif}                  & \cellcolor[HTML]{E0E0E0}                              & 61.0                           & 74.3                           & 98.7                           & 58.2                           & 36.0                           & 58.7                           & 58.4                           & 56.1                           & 99.4                           & 73.6                           & 52.9                           & 74.0                           & 73.8                           & 71.9                           \\
\rowcolor[HTML]{E0E0E0} 
RadarPNv2~\cite{schumann2020tiv}                  & \multirow{-2}{*}{\cellcolor[HTML]{E0E0E0}aggregation} & 61.9                           & 75.0                           & 98.7                           & 63.8                           & \textbf{38.8} & 58.5                           & 51.0                           & 61.0                           & 99.4                           & 77.9                           & \textbf{55.9} & 73.8                           & 67.5                           & 75.8                           \\ \hline
\change{Point Voxel Transformer~\cite{zhang2021arxiv}}                         &                                                       & \change{45.9}                           & \change{57.5}                           & \change{99.3}                           & \change{47.5}                           & \change{7.3}                            & \change{47.5}                           & \change{54.6}                           & \change{19.2}                           & \change{99.6}                           & \change{64.4}                           & \change{13.6}                           & \change{64.4}                           & \change{70.6}                           & \change{32.2}                           \\
Point Transformer~\cite{zhao2021iccv}          &                                                       & 55.6                           & 68.1                           & 99.3                           & 58.1                           & 15.2                           & 56.8                           & 55.1                           & 48.9                           & 99.6                           & 73.5                           & 26.4                           & 72.5                           & 71.1                           & 65.6                           \\
Gaussian Radar Transformer &        \multirow{-3}{*}{single-scan}                                        & \textbf{68.5} & \textbf{79.8} & \textbf{99.4} & \textbf{69.6} & 36.3                           & \textbf{71.2} & \textbf{71.2} & \textbf{62.8} & \textbf{99.7} & \textbf{82.1} & 53.2                           & \textbf{83.2} & \textbf{83.2} & \textbf{77.1}\\ \hline
\end{tabular}%
}
\caption{Semantic segmentation results of moving objects on the RadarScenes test set in terms of IoU and $F_{1}$ scores. The results of RadarPNv1~\cite{schumann2018icif} and RadarPNv2~\cite{schumann2020tiv} are calculated based on the reported confusion matrix.
  \vspace{-0.4cm}}
  \label{tab:resall}
\end{table*}
\section{Implementation Details}
\label{sec:impl}
\change{We construct our architecture based on the self-attention mechanism. The central building blocks are the GTL and the attentive down- and upsampling modules to extract discriminative features for point cloud understanding. The backbone adopts the U-Net architecture of Point Transformer~\cite{zhao2021iccv} with an encoder-decoder architecture including skip connections. First, we directly extract features of the sparse input point cloud by a GTB and increase the per-point feature dimension to 32. The resulting features are progressively down-sampled by four consecutive stages where each reduces the cardinality of the point cloud by a factor of two resulting in [N/2, N/4, N/8, N/16] points. The per-point features are further gradually increased to 64, 128, 256, and 512. The individual stages include the GTB and attentive downsampling modules in the encoder part, which are replaced by attentive upsampling modules in the decoder part of the network. The per-point features maps of the final decoder layer are processed by an MLP with two fully connected layers to obtain point-wise semantic classes $\mathcal{P}^S=\{p^S_1,\dots,p^S_N\}$, where $p^S_i\in \{1, \dots, C\}$.}

\change{We implement the Gaussian Radar Transformer in PyTorch~\cite{paszke2019nips}. To train the network, we utilize the SGD optimizer with an initial learning rate of 0.05, a momentum of 0.9, and a cosine annealing learning rate scheduler~\cite{loshchilov2017iclr}. The batch size $b$ is set to 32. The loss combines the Lovász loss~\cite{berman2018cvpr} and weighted cross-entropy. We follow Schumann~\etalcite{schumann2020tiv} and set the weights of the cross-entropy loss for dynamic objects to 8.0 and for static to 0.5 to account for the class imbalance of the data set. For the attentive sampling operations, we define $k=9$ for the $k$NN operation, and for the Gaussian transformer layer, we restrict the local area to $N_l = 16$. We define $G(x)$ as:
\begin{align}
    G(x) = \exp\left({{\frac{ - x^2 }{2}}}\right),
\end{align}
such that for $x=0$ the attention weight is $G(x)=1$.
Additionally, we apply data augmentation, which includes scaling, rotation around the origin, jitter augmentation of the coordinate features, and instance augmentation.}
\begin{figure*}[t]
  \centering
  \def\svgwidth{\textwidth}
  \fontsize{8pt}{8pt}\selectfont
     \def\svgwidth{\textwidth}
     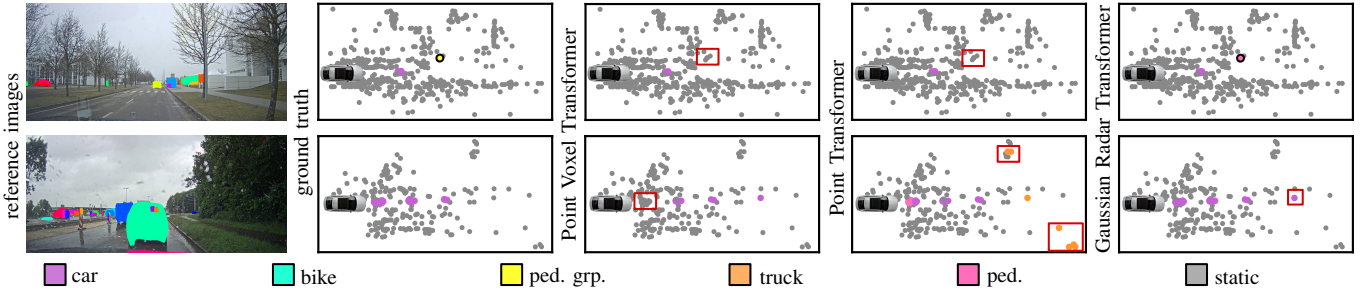
  \caption{\change{Qualitative results of Point Transformer~\cite{zhao2021iccv}, Point Voxel Transformer~\cite{zhang2021arxiv}, and Gaussian Radar Transformer on the test set of RadarScenes~\cite{schumann2021icif}}.}
    \vspace{-0.2cm}
  \label{fig:resw}
  \vspace{-0.4cm}
\end{figure*}
\section{Experimental Evaluation}
\label{sec:exp}

The main focus of this work is to enhance the semantic segmentation of moving objects in sparse and noisy radar point clouds.
We present our experiments to show the capabilities of our method and to support our key claims that our approach achieves state-of-the-art performance in semantic segmentation of moving objects in single-scan radar point clouds without exploration of temporal dependencies or the aggregation of scans. Moreover, we demonstrate that the Gaussian transformer layer and the attentive up- and downsampling modules improve feature extraction and contribute to the final performance. Our fully attentive network is able to extract valuable features from the Doppler velocity and radar cross section provided by the radar sensor.

\subsection{Experimental Setup}
We train and evaluate our method on RadarScenes~\cite{schumann2021icif}, which is the only large-scale, open-source radar data set including point-wise annotations for varying scenarios. The data set consists of 158 annotated sequences. We use the recommended 130 sequences for training and split the remaining 28 sequences into validation (sequences: 6, 42, 58, 85, 99, 122) and test set. The RadarScenes~\cite{schumann2021icif} data set is split into separate scans for each of the four sensors. Since the field-of-view of the sensors is restricted to certain areas, we derive detailed information about the surrounding by merging the individual sensor data from the four sensors into a single radar point cloud. The measurement times and the pose information are given, which enables a transformation into a common coordinate system. We aggregate four scans, one of each sensor, which results in the final input point clouds with transformed local coordinates. To evaluate the performance, Schumann~\etalcite{schumann2021icif} propose the point-wise macro-averaged $F_1$ scores based on all five moving object classes and the static background class ($C=6$). We further report the $IoU=\frac{TP}{TP+FN+FP}$ and $mIoU=\frac{1}{C}\sum_{i=1}^{C}{ IoU_i}$ scores, which are common for semantic segmentation tasks~\cite{behley2019iccv}.
\change{We train each network using its specific hyperparameters with two Nvidia RTX A6000 GPUs over 50 epochs on the training set and report the results on the test set.} For more details on the training regime for Point Transformer\footnote{\url{https://github.com/POSTECH-CVLab/point-Transformer}}, we refer to the original paper~\cite{zhao2021iccv}.

\subsection{Semantic Segmentation of Moving Objects}
The first experiment presents the performance of our approach on the RadarScenes test set to investigate the claim that we achieve state-of-the-art results for semantic segmentation of moving objects in sparse and noisy radar point clouds without the aggregation of scans or the exploration of temporal dependencies. In this experiment, we compare our Gaussian Radar Transformer with the recent and high-performing Point Transformer by Zhao~\etalcite{zhao2021iccv} as well as the baselines provided by Schumann~\etalcite{schumann2018icif,schumann2020tiv}. \change{We selected Point Transformer as a reference since the method meets the following requirements: (1) single-scan input for comparability; (2) point-based method, since the voxelization leads to discretization artifacts and hence a loss of information, see Point Voxel Transformer in \tabref{tab:resall}; (3) very good performance on different benchmarks including semantic scene understanding. Furthermore, the Point Transformer~\cite{zhao2021iccv} utilizes vector attention, which is beneficial for point cloud understanding.}

Our Gaussian Radar Transformer outperforms the existing
methods in terms of both, mIoU and $F_1$ score, as displayed in \tabref{tab:resall}. Especially, we achieve superior performance on five of the six classes, except pedestrian. \change{We assume that the individual detection in radar scans contains important information, which is why strict point-based methods enhance the performance compared to Point Voxel Transformer.}
The baselines exploit temporal dependencies of consecutive radar scans within a memory feature map, utilize additional global coordinates or densify the point clouds by aggregation. The exact comparison of the results is difficult because Schumann \etal work on a subset of the officially released data set. However, the IoU for the class pedestrian indicates that the exploration of temporal information is beneficial for small instances. We suspect that the consistent detection of pedestrians over the whole sequence, which is difficult for strict single-scan approaches, further improves the performance. Nevertheless, the Gaussian Radar Transformer considerably improves the IoU for the class pedestrian as opposed to Point Transformer by more than 19 absolute percentage points. \figref{fig:resw} shows some qualitative results \change{on the test set}. Notably, our approach achieves superior performance under adverse weather including rain and fog. 
\subsection{Ablation Studies on Method Components}
The first ablation study presented in this section is designed to support our second claim that our proposed self-attention modules each contribute to the advancements of the Gaussian Radar Transformer. To assess the influence of the different components of our fully attentive backbone, we evaluate the performance in terms of mIoU and $F_1$ score on the validation set. To replace our proposed modules, we follow commonly used network designs. We substitute the Gaussian function by the softmax function and keep the rest of the Gaussian transformer layer as it is. For the attentive downsampling, we utilize local max pooling and we exchange attentive upsampling by trilinear interpolation based on an inverse distance weighted average. 
\tabref{tab:components} summarizes the influence of different components on the performance in terms of mIoU on the validation set.

In configuration (A), we replace each module by its substitute, which leads to a noticeable decrease in mIoU\@. We suspect that the commonly used modules are highly optimized for denser point clouds but struggle to capture fine-grained information from sparse and noisy radar point clouds.
In (B), we add attentive downsampling~(ADS), see \secref{sec:atdown}, 
which introduces a smooth information exchange within the downsampling step of individual points, visibly improving the results.
In (C), we add the attentive upsampling~(AUS) module to enlarge the receptive field and include encoded feature information to optimize the information aggregation, see \secref{sec:atup}. The larger receptive field resulting from the increased local area from three (trilinear) to nine points improves the $F_1$ score by 3.3 and the mIoU by 4.5 absolute percentage points. Although the AUS only affects the features of the decoder part it leads to an additional improvement of mIoU by 0.8 absolute percentage points compared to AUS in (B).
In (D), we add the attentive up- and downsampling which further enhance the performance. This shows the importance of the attentive sampling modules for sparse radar point cloud processing.
In (E), we utilize the fully attentive network to illustrate the improvement due to the usage of the Gaussian function by decoupling individual points, see \secref{sec:gtl}, resulting in the best performance. In conclusion, the Gaussian function and the attentive up- and downsampling are essential to extract valuable features from sparse and noisy radar point clouds. 
\begin{table}[t]
  \centering
  \begin{tabular}{c c c c c c c}
    \toprule
    \# & ADS  &AUS     & GTL   & $F_1$ & mIoU \\
    \midrule
    A  &   &            &              &    74.0        & 61.0           \\
    
    B  & \checkmark &   &                &  77.0       & 64.7             \\
    C  &            &    \checkmark         &    &   77.3        & 65.5             \\

    D  & \checkmark &  \checkmark          &   &     78.8        & 66.8           \\
    E  & \checkmark & \checkmark & \checkmark   &  \textbf{79.4}    & \textbf{68.3}             \\
    \bottomrule
  \end{tabular}
  \caption{Influence of the different components of the approach in terms of mIoU and $F_1$ score on the RadarScenes validation set.\vspace{-0.2cm}}
  \label{tab:components}
  \vspace{-0.3cm}
\end{table}
\subsection{Ablation Studies on Input Features}
The third experiment evaluates the performance depending on the provided information by the radar sensor and demonstrates that our approach is capable of capturing complex local structures within the features to enhance mIoU\@. For this experiment, we utilize our Gaussian Radar Transformer and add to the position information of $x$ and $y$ coordinates, the ego-motion compensated Doppler velocity $v$, the radar cross section $\sigma$, or both. \tabref{tab:features} displays the influence of the input features $\mathbf{x}^F$ on the validation set performance. As we presume, the ego-motion compensated Doppler velocity is especially valuable for semantic segmentation of moving objects since the feature inherently distinguishes between moving and non-moving parts of the environment resulting in an increase of mIoU of 18.2 absolute percentage points. Moreover, we further improve the mIoU if we add the radar cross section features $\sigma$ suggesting that our approach extracts valuable features for the downstream task from additional sensor information. Hence, the Gaussian Radar Transformer achieves the best performance including radar cross section and ego-motion compensated Doppler velocity.

In summary, our evaluation supports our statement that our method provides competitive semantic segmentation performance of moving objects in single-scan, sparse radar point clouds. At the same time, our method exploits self-attention modules which enhance the performance in multi-dimensional radar data processing outperforming state-of-the-art approaches. Thus, we support all our claims with this experimental evaluation.

\begin{table}[t]
  \centering
  \begin{tabular}{c  c c c}
    \toprule
     Input Features    & $F_1$     & mIoU \\
    \midrule
    $x^F=(x,y)$   &                 56.0                  & 43.7         \\

    $x^F=(x,y,\sigma)$              &    63.7                      & 50.1             \\
    $x^F=(x,y,v)$  &                    75.0      & 62.0            \\

      $x^F=(x,y,v,\sigma)$   &  \textbf{79.4}    & \textbf{68.3}           \\
    
    \bottomrule
  \end{tabular}
  \caption{Influence of the different input features in terms of mIoU and $F_1$ score on the RadarScenes validation set.\vspace{-0.2cm}}
  \label{tab:features}
  \vspace{-0.3cm}
\end{table}

\section{Conclusion}
\label{sec:conclusion}

In this paper, we presented a novel approach to perform semantic segmentation of moving objects in sparse, noisy, single-scan radar point clouds obtained from automotive radars. Our method exploits the self-attention mechanism throughout the network and replaces the softmax normalization of the transformer by a Gaussian. This allows us to successfully segment moving objects and improve the feature extraction by decoupling individual points. 
We implemented and evaluated our approach on the RadarScenes data set, providing comparisons to other methods and supporting all claims made in this paper. The experiments suggest that the proposed architecture achieves good performance on semantic segmentation of moving objects. We assessed the different parts of our approach and compared them to other existing techniques.
Overall, our approach outperforms the state of the art both in $F_1$ score and mIoU, taking a step forward towards sensor redundancy for semantic segmentation for autonomous robots and vehicles.

\bibliographystyle{plain_abbrv}

\bibliography{glorified,new}

\end{document}